\documentclass[letterpaper]{article}
\usepackage{aaai25}
\usepackage{times}  
\usepackage{helvet}  
\usepackage{amsfonts} 
\usepackage{courier}  
\usepackage[table]{xcolor}
\usepackage{booktabs} 
\usepackage{comment}
\usepackage{amsmath} 
\usepackage[hyphens]{url}  
\usepackage{graphicx} 
\usepackage{natbib}  
\usepackage{caption} 
\usepackage{algorithm}
\usepackage{algorithmic}
\usepackage{enumitem}

\usepackage{newfloat}
\usepackage{listings}
\DeclareCaptionStyle{ruled}{labelfont=normalfont,labelsep=colon,strut=off} 
\floatstyle{ruled}
\newfloat{listing}{tb}{lst}{}
\floatname{listing}{Listing}
\title{WaveHiT-SR: Hierarchical Wavelet Network for Efficient Image Super-Resolution}
\author {
    Fayaz Ali\textsuperscript{\rm 1},
    Muhammad Zawish\textsuperscript{\rm 2},
    Steven Davy\textsuperscript{\rm 2},
    Radu Timofte\textsuperscript{\rm 1}
}
\affiliations {
    \textsuperscript{\rm 1}Computer Vision Lab, CAIDAS \& IFI, University of Würzburg, Germany\\
    \textsuperscript{\rm 2}Technological University Dublin, Ireland\\

}

\usepackage{bibentry}

\begin{document}

\maketitle

\begin{abstract}
Transformers have demonstrated promising performance in computer vision tasks, including image super-resolution (SR). The quadratic computational complexity of window self-attention mechanisms in many transformer-based SR methods forces the use of small, fixed windows, limiting the receptive field.  In this paper, we propose a new approach by embedding the wavelet transform within a hierarchical transformer framework, called (WaveHiT-SR). First, using adaptive hierarchical windows instead of static small windows allows to capture features across different levels and greatly improve the ability to model long-range dependencies. Secondly, the proposed model utilizes wavelet transforms to decompose images into multiple frequency subbands, allowing the network to focus on both global and local features while preserving structural details. By progressively reconstructing high-resolution images through hierarchical processing, the network reduces computational complexity without sacrificing performance. The multi-level decomposition strategy enables the network to capture fine-grained information in low-frequency components while enhancing high-frequency textures. Through extensive experimentation, we confirm the effectiveness and efficiency of our WaveHiT-SR. Our refined versions of SwinIR-Light, SwinIR-NG, and SRFormer-Light deliver cutting-edge SR results, achieving higher efficiency with fewer parameters, lower FLOPs, and faster speeds.
\end{abstract}

%
\section{Introduction}
\label{sec:intro}
Several downstream computer vision tasks require image enhancement to aid in accurate decision-making. Single-image super-resolution (SISR) is one of the prominent tasks for image enhancement. SISR is commonly adopted in plethora of usecases ranging from remote sensing \cite{dharejo2021twist}, medical imaging \cite{dharejo2022multimodal} to facial recognition \cite{wang2024structure}. It aids the decision making process by providing a high resolution (HR) output with improved perceptual quality from a low resolution (LR) input \cite{alexey2020image, lim2017enhanced, ahn2018fast, bittner2023lstm}. Despite its wide adoption, the task of SISR is inherently challenging because of its ill-posed nature. This drives the researchers in vision community to propose efficient but at the same time robust models for this low-level vision task.

The deep convolutional neural network (CNN) gained immense popularity among early methods to solve SISR. CNNs were attractive thanks to their effective architecture for retaining and generating fine grained details in images. This resulted in a rapid transition from classical statical approaches \cite{lee2017unraveling} to learning-based methods \cite{guo2017deep, zhang2018residual, zhang2019residual, dong2014learning} for improved SISR. Regardless of this, existing CNN based works heavily incorporate spatially invariant kernels with an aim to extract only local features. This hampers the adaptability of CNNs to observe and model pixel-level relationships, which is important for enhanced perceptual quality. In addition, researchers focus on making CNNs more deeper and complex to improve SR performance \cite{dong2014learning, lim2017enhanced}. However, it is observed that depth does not contribute to enhancement but overlooks vital information such as high frequency details of certain objects due to large number of downsampling operations 
\begin{figure}[t]
    \centering
    \includegraphics[width=0.48\textwidth]{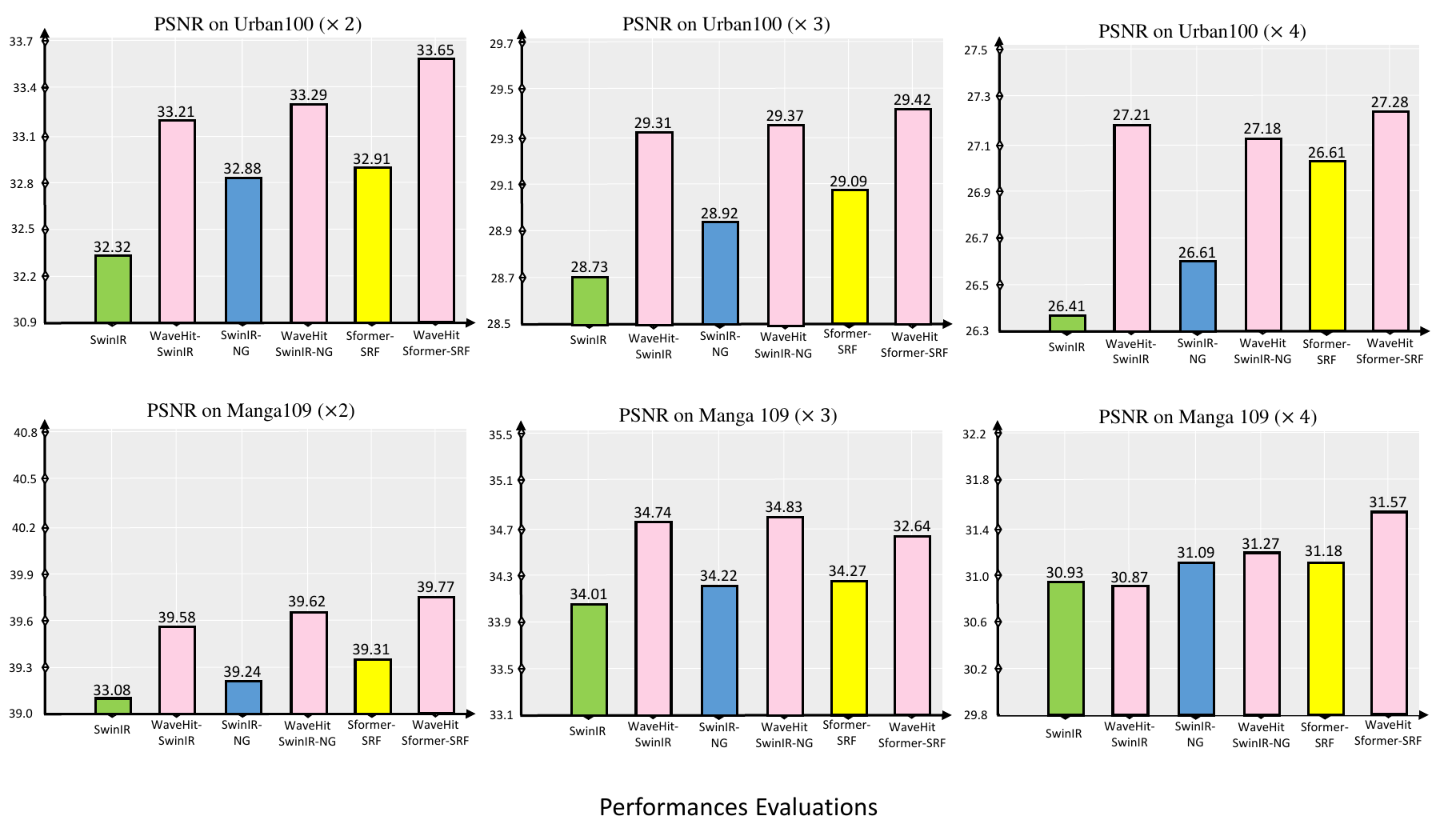}
    \caption{Comparison of efficient SR transformers: SwinIR-Light (SIR) \cite{liang2021swinir}, SwinIR-NG (SNG) \cite{choi2023n}, and SRFormer-Light (SRF) \cite{zhou2023srformer} and their WaveHiT-SR counterparts (WaveHiT-SIR, WaveHiT-SNG, WaveHiT-SRF) on the Urban100 and Manga109 benchmarks.}
    \label{fig:1}
\end{figure}

Recently, new architectures based on transformers offer great utility for tasks beyond natural language processing (NLP) \cite{vaswani2017attention,zhu2020deformable}. Their ability to perform self-attention mechanism on given features has received enormous attention for various high \cite{yuan2021hrformer, wang2021pyramid, li2023uniformer, zheng2021rethinking} and low level vision tasks \cite{liang2021swinir, li2021efficient, cao2021video, wang2022uformer, chen2023activating, wang2023omni, zhou2023srformer}. Several vision transformers (ViTs) have been proposed based on sliding window and self-attention mechanisms that extend the legacy of swin transformers \cite{liang2021swinir, zhang2022swinfir}. We observe a common trend among such architectures is their inability to capture long-range dependencies without increasing additional computing complexity \cite{wang2021pyramid, liang2021swinir, zhang2022swinfir}. Moreover, the self-attention mechanism, despite being powerful, leads to weak cross-window interaction within ViTs such as in SwinIR \cite{liang2021swinir}. To address these challenges, this work introduces a hybrid attention block that uses channel and wavelet attention. The wavelet block aids in reconstructing high frequency details within objects, while channel attention boosts the utilization of global information for efficient and robust SR. In addition, inspired by the importance of multiscale feature representation for SR, we also propose hierarchical windows in transformer layers. These windows allow ViT to capture multiscale features through moderately expanding receptive fields. Inspired by the advantages of multiscale feature integration \cite{hui2019lightweight,kim2016accurate, lai2017deep}, we offer a new approach that restructures common transformer-based SR networks into hierarchical transformers for more efficient and accurate image SR (WaveHiT-SR). In summary, the WaveHiT-SR framework consists of three key contributions:

\begin{itemize}
\item  We introduce WaveHiT-SR, an effective approach that adapts popular transformer-based SR methods into a hierarchical transformer framework, enhancing SR performance by leveraging multiscale features and capturing long-range dependencies. A comparison with state-of-the-art is shown in Figure \ref{fig:1}. 

\item We propose a WaveAttention mechanism within hierarchical transformers to capture texture and edge details efficiently, achieving linear complexity with respect to window size and enabling the use of large hierarchical windows, such as $64 \times 64$

\item We reframe SwinIR-Light  (SIR) \cite{liang2021swinir}, SwinIR-NG (SNG) \cite{choi2023n}, and SRFormer-Light (SRF) \cite{zhou2023srformer} as WaveHiT-SR models specifically, \textit{\underline{WaveHiT-SIR}}, \textit{\underline{WaveHiT-SNG}}, and \textit{\underline{WaveHiT-SRF}}  achieving enhanced performance with a reduced number of parameters, decreased FLOPs, and faster runtime
\end{itemize}

\begin{figure*}[t]
\centering
\includegraphics[width=0.9\textwidth]{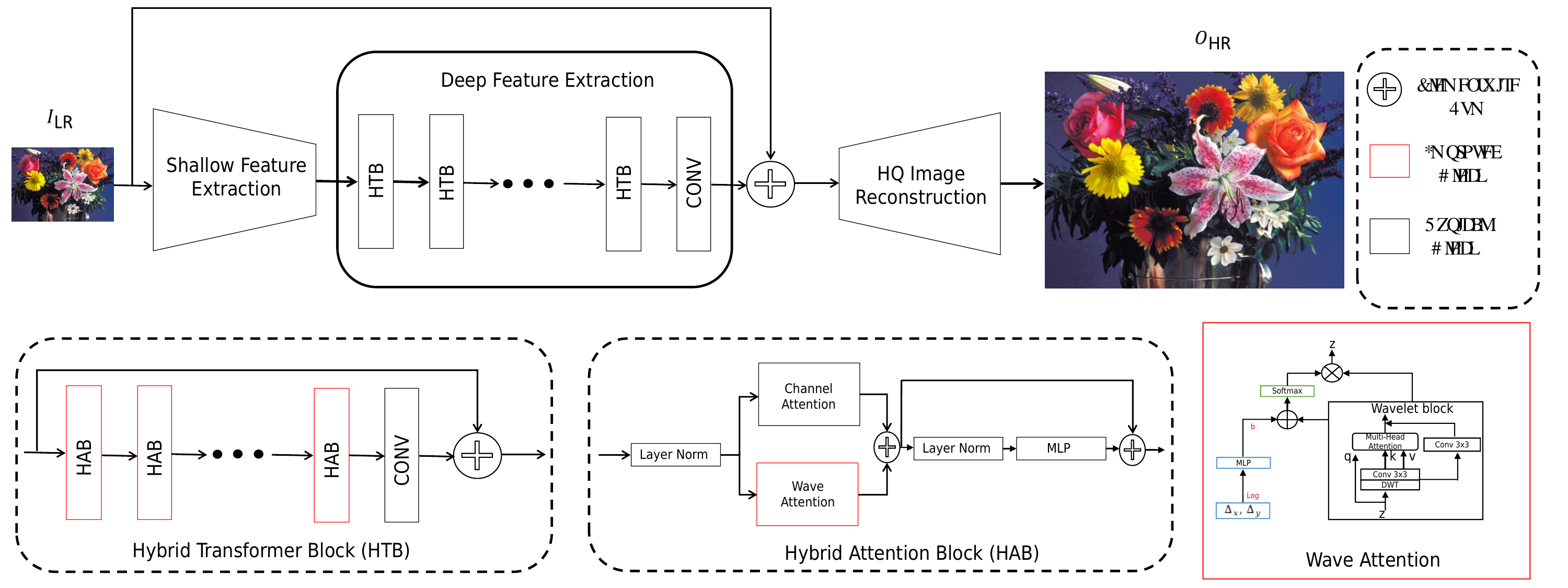}
\caption{Overview of the proposed SR framework, with specific block- and layer-level enhancements from our WaveHiT-SR highlighted in red.}
\label{fig2}
\end{figure*}

\section{Related Work}
\label{sec:Related Work}
A plethora of deep neural networks have been proposed to address the prevalent challenge of SR in the past decade.
\subsection{SR with CNNs}
\label{sec:CNN}
CNNs were among early adopters for the task of SISR thanks to their remarkable performance and parametric efficiency. Among others, SRCNN \cite{dong2014learning} laid the foundation of CNNs being applied for the SISR by learning a non-linear function to map a  bicubic interpolation based LR image to a HR output via a few convolutional layers. Inspired from this, a followup study by Kim et al. \cite{kim2016accurate} involved a deeper VGG-19 with residual learning capacity to improve SR performance. Advanced models like EDSR \cite{lim2017enhanced} topped the NTIRE2017 challenge \cite{agustsson2017ntire} by proposing residual blocks but skipping batch normalization layers for efficiency and accuracy. Recent works with CNNs attempted to develop more deeper and complex architectures. For example, MemNet \cite{tai2017memnet} and RDN \cite{zhang2018residual} exploited dense blocks to leverage intermediate features from individual layers. Different from these works, RCAN \cite{zhang2019residual} and VDSR \cite{zhang2019residual} made great efforts to capture high-frequency details via residual structures. Inspired from the intuition that high-frequency details are important for improved SR, this work also aims to preserve such details by introducing wavelet based attention. 

\subsection{SR with Vision Transformers}
\label{sec:ViT}
Inspired from the natural success of transformers in NLP \cite{vaswani2017attention}, several researchers incorporated ViTs for both high-level vision \cite{yuan2021hrformer, wang2021pyramid, li2023uniformer} and low level vision tasks such as SISR \cite{liang2021swinir, li2021efficient, cao2021video, wang2022uformer, chen2023activating, wang2023omni, zhou2023srformer}. Building on the success of self-attention (SA) mechanism, SwinIR \cite{liang2021swinir} is a prominent work which introduced shifted window self-attention for robust image restoration. As an extension of this, several works emerged such as omni self-attention (OSA) \cite{wang2023omni}, and permuted self-attention (PSA) \cite{zhou2023srformer} for more complex, multi-scale feature extraction through custom self attention modules. However, SwinIR and similar approaches compute SA on static, constant size windows overlooking the long-range feature dependency. In contrast, UFormer \cite{wang2022uformer} claims to capture local and global dependencies effectively by proposing a multi-scale restoration module. Following this, SRFormer based on OSA \cite{wang2023omni} strikes the balance between channel and spatial features by feeding an optimized SA mechanism for SR. Regardless of several such improvements, a common gap still exists in leveraging hierarchical features at multi scales with ViTs, which is filled by this work to boost the SR performance.

\subsection{Frequency Domain for SR}
\label{sec:Frequecny SR}
Recent works have incorporated the low and high-frequency image features simultaneously for improved SISR. For example, authors in \cite{baek2020single} leveraged multi-branch CNN architectures to decompose feature set into frequency in fourier domain. Similarly, fourier-driven architectures based on ViTs were introduced such as SwinFIR \cite{zhang2022swinfir} exploits fast Fourier convolution to incorporate image-wide receptive fields. Despite such innovations, fourier-based models tends to overlook the high-frequency features, limiting their SR performance gains. In contrast, several works such as such as DWSR \cite{guo2017deep}, Wavelet-SRNet \cite{huang2017wavelet}, PDASR \cite{zhang2022perception} leverage wavelet transforms in SR to address such limitations. For example, authors in \cite{zhang2022perception} improve the visual fidelity and avoids the artifacts by conditioning the reconstruction on low-level wavelet coefficients. Latest work from WGSR \cite{korkmaz2024training} replaced the traditional $L_1$ loss with a weighted combination of wavelet losses to efficiently eliminate the artifacts and improve the SR quality eventually. Inspired from the success of wavelets, this work introduces a hybrid attention block in ViT incorporating wave attention for multiscale and efficient SR. 
\vspace{-10pt}

\section{Method}
\label{sec:Method}
The foundational approach of WaveHiT-SR, described in the section on the Hierarchical Transformer, includes the Hybrid Attention Block (HAB), Layer-Based Architecture, and Dual Feature Extraction (DFE), with a particular emphasis on the layer-based design.

\subsection{Hierarchical Transformer}
\label{sec:HiT}
\begin{figure*}
  \centering
  \includegraphics [height=9cm, width=15cm]{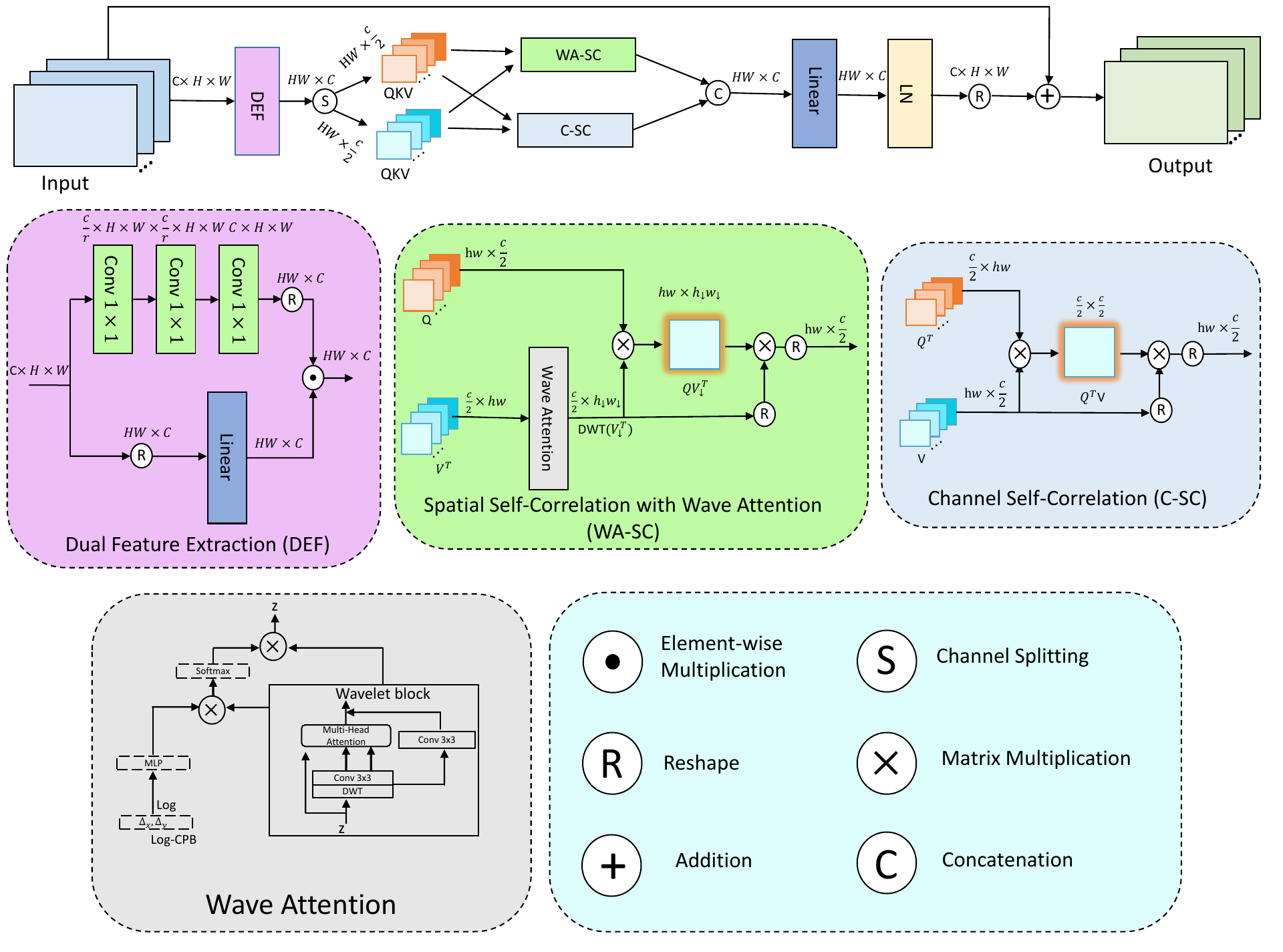}
  \caption{The WaveHiT-SR framework’s layer-level design includes dual feature extraction (DFE) and the use of wave attention and channel self-correlation (WA-SC and C-SC). DFE focuses on spatial and channel feature extraction, while WA-SC and C-SC provide efficient hierarchical aggregation. Here, WA-SC replaces traditional spatial self-correlation, leveraging Discrete Wavelet Transform (DWT) to capture frequency-based correlations, enhancing feature richness and maintaining linear computational complexity relative to window sizes.}
  \label{Fig4}
\end{figure*}

First, we analyze the conventional structure typically used in transformer-based SR techniques \cite{choi2023n, liang2021swinir, zhou2023srformer}. As illustrated in Figure \ref{fig2}, these frameworks generally include an initial convolutional layer that extracts shallow features $F_{S}\in \mathbb{R} ^ {C\times H\times W} $ from the low-resolution input $I_{LR}\in \mathbb{R} ^ {3\times H\times W} $, a feature extraction stage using transformer blocks (TBs) to capture deep features $F_{D}\in \mathbb{R} ^ {C\times H\times W} $, and a reconstruction module that produces the high-resolution output $I_{HR}\in \mathbb{R} ^ {3\times sH\times sW} $, with $s$ as the scaling factor. The feature extraction module comprises transformer blocks (TBs) with cascaded transformer layers (TLs) and convolutional layers. Each TL includes components for self-attention (SA), a feed-forward network (FFN), and layer normalization (LN). To alleviate the quadratic computational load of SA on large inputs \cite{alexey2020image}, window partitioning is commonly applied in TLs, restricting SA to local regions through a technique known as window self-attention (W-SA) \cite{liang2021swinir, liu2021swin}.
To aggregate hierarchical features more effectively, we redesign the SR framework using hierarchical transformers enhanced with DWT (Discrete Wavelet Transform). As depicted in Figure \ref{fig2} and \textbf{Figure (DWT: Supplementary file)}, this approach involves applying DWT to downsample keys and values, which amplifies the impact of self-attention in feature learning by enabling more focused, noise-reduced attention across multiple scales. As depicted in Figure \ref{Fig4}, this approach involves: (i) block-level hierarchical windows to enhance long-range and multi-scale information capture, (ii) a Wave Attention module to improve texture detail, and (iii) an efficient window self-attention, adapted from HiT-SR \cite{aslahishahri2024hitsr}, that reduces computational costs with large windows.

\subsection{Hybrid Attention Block (HAB)}
\label{sec:HAT}
We introduce a Hybrid Attention Block (HAB) that combines channel attention and wave attention to enhance both local and global feature extraction. As shown in Figure \ref{fig2}, channel attention leverages global information to dynamically adjust attention weights across channels, selectively activating pixels and focusing on key features across multiple scales. To capture long-range dependencies and multi-scale information, the HAB incorporates wave attention, expanding spatial connections to capture detailed structures essential for high-resolution super-resolution outputs. This combined attention approach ensures fine detail retention across spatial regions. For the multi-head self-attention mechanism (illustrated in Figure \ref{fig2} and  \textbf{Figure (DWT: Supplementary file)} (a)), the input 2D feature map $X \in \mathbb{R} ^ {H\times W\times D}$, where $H$, $W$, and $D$ denote height, width, and channel count, respectively, is reshaped into a sequence of $n = H \times W$ patches, with D representing the dimensionality of each patch. The sequence \textbf{X} is then split into three parallel groups: queries \textbf{(Q)}, Keys \textbf{(K)}, and Values \textbf{(V)} where $Q \in \mathbb{R} ^ {n \times D}$, keys $K \in \mathbb{R} ^ {n \times D}$, and values $V \in \mathbb{R}^{n \times D}$. The multi-head self-attention module partitions each \textbf{(Q)}, Keys \textbf{(K)}, and Values \textbf{(V)} into $N_{h}$  segments along the channel dimension, yielding queries
$Q_{j} \in \mathbb{R}^{n \times D_{h}}$ , keys $K_{j}  \in \mathbb{R}^{n \times D_{h}}$ , and values $V_{j} \in \mathbb{R}^{n \times D_{h}}$ for the $j^{th}$. Existing multi-scale ViT backbones reduce computation for keys and values through average pooling \cite{wang2022pvt} and kernel pooling \cite{fan2021multiscale}. In our approach, as illustrated in Figure \ref{Fig4}, we replace spatial attention with wave attention, leveraging DWT for downsampling by a factor of $\gamma=2$. This reduces the spatial dimensions of the input feature map $X$ by half using the down-sampling operator. 

\subsection{Layer-Based Architecture}
\label{sec:Layer Arch}

To replace the Layer-Level design in HiT-SR \cite{aslahishahri2024hitsr} with wavelet-based techniques, we can redefine the key components of HiT-SR's architecture using wavelet transforms. This approach will help achieve multi-level feature extraction, spatial and channel self-correlation \textit{(\textbf{See Supplementary File})}, and efficient aggregation of hierarchical information. Below is a step-by-step breakdown of the methodology:

\subsection{Dual Feature Extraction (DFE)} 
Given the input feature map 
$F \in \mathbb{R}^{H \times W \times C}$
  from the previous layer, where $H $ and 
$W$ denotes spatial dimensions, and 
 $C $ is the channel count, we apply the DWT to decompose $F $ into frequency-specific sub-bands. This yields both low- and high-frequency components that capture different levels of detail and spatial structures: 
\begin{equation}
    F_{\text{wavelet}} = \text{DWT}(F) = \{ F_{HH}, F_{HL},F_{LH}, F_{LL}  \}
\end{equation}
Where 
$F_{HH} $ and $F_{HL}$
  represents low-frequency components that capture global structure, and 
$F_{LH}$ and $ F_{LL}$
  represents high-frequency components that capture finer details and textures.
In contrast to the traditional Dual Feature Extraction (DFE) approach in HiT-SR, we use the Discrete Wavelet Transform (DWT) to break the image into frequency sub-bands. 

\begin{equation}
    \text{DFE}(X) = X_{\text{ch}} \odot X_{\text{wave}} ,
\end{equation}
\begin{align*}
    X_{\text{ch}} &= \text{Linear}(X), \\
    X_{\text{wave}} &= \text{WaveConv}(X) .
\end{align*}

Here, 
$X_{\text{ch}}$
  represents the channel attention from a linear transformation of 
$X$, while $ X_{\text{wave}}$ introduces wave attention through a wave-based convolution operation.

The notation 
$\odot$ refers to element-wise multiplication. The reshaped channel feature 
$X_{\text{ch}} \in \mathbb{R}^{H \times W \times C}$
  and spatial feature 
 $ X_{\text{wave}}  \in \mathbb{R}^{H \times W \times C}$
  are extracted using linear layers and wavelet convolutional layers, respectively. In the spatial branch, we apply a wavelet transform to break the image into multiple frequency sub-bands. Instead of a traditional attention mechanism, wavelet attention is applied using the DWT, capturing both low- and high-frequency components efficiently. The multi-resolution features are aggregated by wavelet-based interactions, where spatial and channel features combine in the wavelet domain to form the DFE output. The queries 
$Q \in \mathbb{R}^{H \times W \times \frac{C}{2}}$
  are derived from the DFE output by splitting it, as depicted in Figure \ref{Fig4}

\begin{equation}
    [Q, V] = \text{DFE}(X),
\end{equation}
Next, we partition queries and keys into non-overlapping windows based on the assigned window size (e.g.,$Q_i, V_i \in \mathbb{R}^{h_i w_i \times \frac{C}{2}}$ for the $
i-th $ TL)

\textbf {Wave Attention:} Unlike W-SA, our Wave Attention (WA) efficiently aggregates spatial information by adaptively summarizing the values 
$V_i $ across TLs, using linear layers on the wavelet-transformed components, as detailed in the supplementary material. The extracted features can then be processed separately or combined to ensure rich multi-channel representations.
 Instead of directly applying a linear transformation
$V^{T}_{i,\downarrow} = \text{S-Linear}(V^T_i)$ we can employ wavelet decomposition to process the feature, The DWT decomposes the input feature into multiple frequency sub-bands, capturing both coarse and fine details through low- and high-frequency components.

\begin{equation}
   V^{T}_{i,\downarrow} = \text{DWT}(V^T_i)
\end{equation}
This step captures the multi-resolution information from the input feature $V^T_i$, breaking it into approximation and detail components. where $( V_{\downarrow,i} \in \mathbb{R}^{h_{\downarrow} w_{\downarrow} \times \frac{C}{2}} )$ denotes the projected values with:
\begin{itemize}
    \item If $\alpha_i \leq 1$,  then downsampling keeps the resolution as $[h_{\downarrow}, w_{\downarrow}] = [h_i, w_i]$

    \item If $\alpha_i > 1$ the resolution is modified to $[h_B, w_B]$
  where 
$h_B$
  and 
$w_B$
  denote the adjusted downsampled dimensions.
\end{itemize}
\textbf{Downsampling: }After applying the wave attention (WA), the low-frequency (approximation) components are downsampled to reduce spatial resolution while preserving important hierarchical information. We define the resolution as:

\begin{equation}
    \text{WA-SC}(Q_i, V_{\downarrow,i}) = \frac{Q_i \cdot \text{WA} (V^{T}_{i,\downarrow})}{D} + B \cdot \text{WA}(V^{T}_{i,\downarrow})
\end{equation}

\begin{figure*}[ht]
  \centering
  \includegraphics[width=0.7\linewidth]{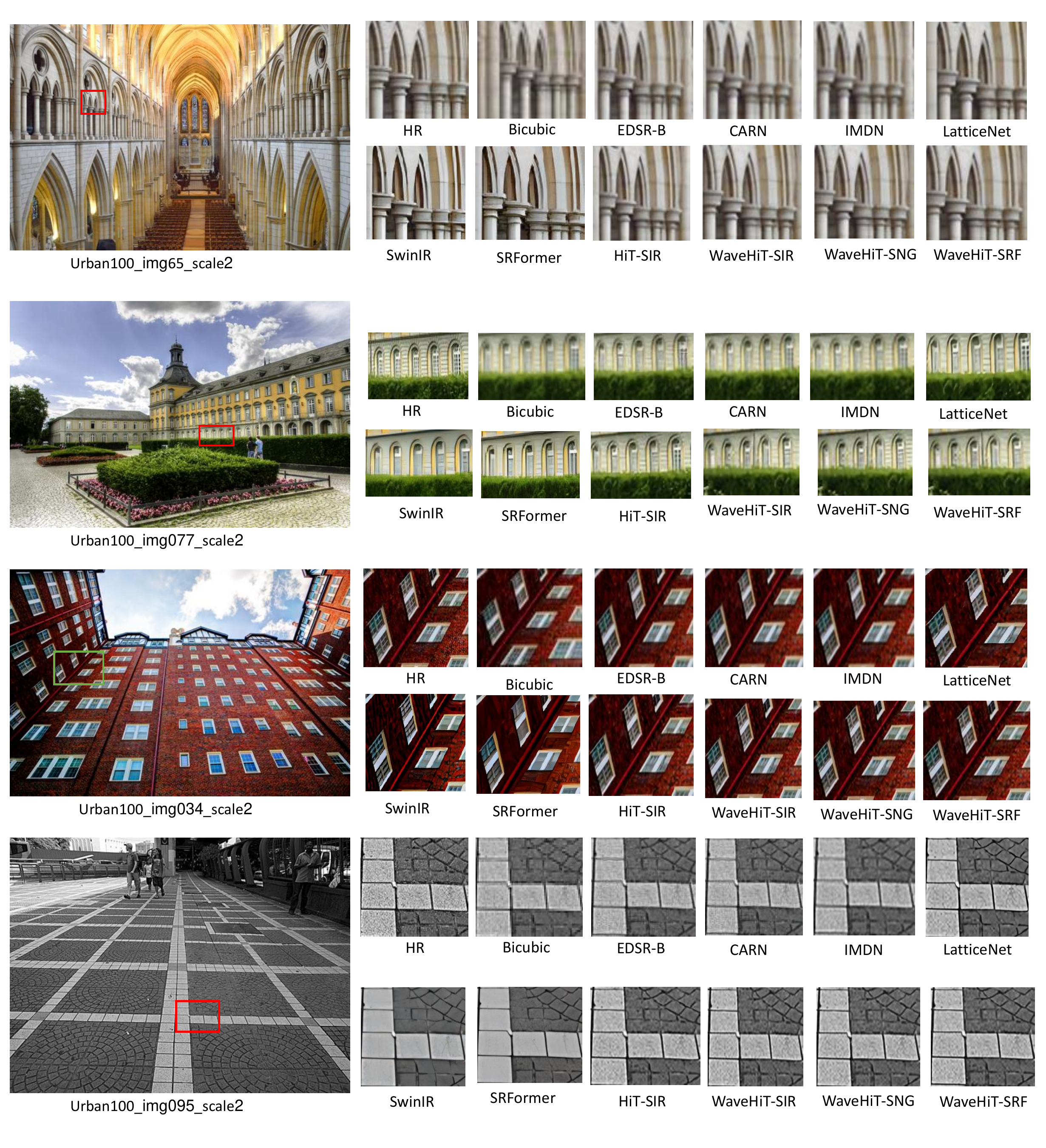}
  \caption{Visual comparisons with state-of-the-art SISR methods on the Urban100 Dataset on Scale ($\times 2$). }
  \label{fig:Urban100}
\end{figure*}

In this version, 
$V^{T}_{i,\downarrow}$
  is replaced with 
$\text{WA} (V^{T}_{i,\downarrow})$ to indicate wavelet-transformed values, the equation name is updated to ``WA-SC" for DWT-based self-correlation, and normalization with 
$D$
 and position encoding 
$B$ are retained as in the original.

Compared to standard W-SA, our WA-SC is more efficient and less complex: (i) Using correlation maps instead of attention maps removes the hardware-heavy softmax, improving inference speed \cite{cai2023efficientvit}; (ii) It supports large windows with linear complexity scaling to window size, as shown in \ref{tab:Complexity}. For an input with \( N \) windows in \( \mathbb{R}^{hw \times C} \), the multiply-add operation counts for W-SA and WA-SC are as follows:

\small
\begin{equation}
\begin{split}
    \text{Mult-Add (W-SA)} &= 2NC(hw)^2,\\
\text{Mult-Add (WA-SC)} &=  2N \cdot C_h \cdot \frac{w}{2} \cdot \frac{h}{2}, 
\label{Equ3}
\end{split}
\end{equation}
\normalsize

Applying DWT reduces the spatial dimensions (width and height) by half, which in turn reduces the computational cost associated with the spatial self-correlation by a factor of 4 for these dimensions combined (i.e., $\frac{w}{2} \cdot \frac{h}{2}=\frac{wh}{4})$.\\
\textit{\textbf{Note: We have provided in-depth information on Spatial and Channel Self-Correlation and Hierarchical Information Aggregation in the supplementary material along with Ablation Study.}}

\


\begin{table}[t]
  \centering
  \setlength{\tabcolsep}{2pt} 
  \small 
  \begin{tabular}{|l|c|c|p{2cm}|}
    \hline
    \rowcolor{cyan!20} \textbf{Attention Type} & \textbf{Complexity} & \textbf{Focus} & \textbf{Summary} \\
    \hline
    Global Self-Attn & $\mathcal{O}((h \cdot w)^2 \cdot C)$ & Quadratic & Scales with all pixel pairs \\
    Window Self-Attn & $\mathcal{O}(N \cdot (h \cdot w)^2 \cdot C)$ & Local & Quadratic within windows \\
    \rowcolor{pink} \textbf{WaveAttention} & $\mathcal{O}(h \cdot w \cdot \log(h \cdot w))$ & Efficient & Near-linear via wavelet transform \\
    \hline
  \end{tabular}
  \caption{Comparison of attention mechanisms in terms of complexity and spatial focus.}
  \label{tab:Complexity}
\end{table}

\section{
Experimental Settings}

We employ the WaveHiT-SR strategy with prominent SR models such as SwinIR-Light \cite{liang2021swinir}, SwinIR-NG \cite{choi2023n}, and SRFormer-Light \cite{zhou2023srformer}, referring to these versions as WaveHiT-SIR, WaveHiT-SNG, and WaveHiT-SRF. To provide a fair assessment, we adjust each method to the HiT-SIR\cite{aslahishahri2024hitsr} version with minimal changes and maintain uniform hyperparameters across all SR transformers. Specifically, we adopt the original SwinIR-Light settings \cite{liang2021swinir} and set the TB number, the TL number, the channel number, and the head number to 4, 6, 60, and 6. The size of the base window is $8 \times 8$, and the hierarchical ratios follow the same structure as HiT-SR.
The training process for HiT-SIR \cite{liang2021swinir}, HiT-SNG \cite{choi2023n}, and HiT-SRF \cite{zhou2023srformer} follows a consistent approach. All models are implemented in PyTorch \cite{paszke2019pytorch} and trained with 64×64 patches and a batch size of 64 for 500K iterations. We optimize the models using L1 loss and the Adam optimizer \cite{lai2017deep} $(\beta_1 = 0.9, \beta_2 = 0.99)$, starting with a learning rate of $6\times 10^{-4}$, which is halved in iterations [250K, 400K, 450K, 475K]. \\
\textbf{Benchmarking:} In line with previous approaches \cite{choi2023n,liang2021swinir, zhou2023srformer}, we train our model on the widely-used DIV2K \cite{agustsson2017ntire} dataset and evaluate on five classic benchmarks: Set5 \cite{bevilacqua2012low}, Set14 \cite{zeyde2012single}, B100 \cite{martin2001database}, Urban100 \cite{huang2015single}, and Manga109 \cite{matsui2017sketch}. Low-resolution images are derived via bicubic degradation from high-resolution images. We evaluate performance at ×2, ×3, and ×4 upscaling levels, using PSNR and SSIM metrics computed on the Y channel in the YCbCr color space.

\begin{table*}
\label{Tab2}
\centering

\caption{Quantitative performance of Super-Resolution methods is evaluated on benchmark datasets using PSNR and SSIM scores. In the results tables, top-performing values are displayed in blue to easily distinguish performance rankings.}
\label{tab:PSNR and SSIM}
\scalebox{0.65}{
\begin{tabular}{lcc|c|c|c|c|c|c}
\toprule
 \rowcolor[HTML]{D3D3D3}\textbf{Method}    & \multicolumn{2}{c|}{\textbf{Complexity}} & \textbf{Scale} & \multicolumn{5}{c}{\textbf{Benchmark Datasets}} \\ 

 &\textbf{Param. Count} & \textbf{ FLOPs} & \textbf{Scale} & \textbf{Set5}  & \textbf{ Set14} & \textbf{ B100} & \textbf{ Urban100} & \textbf{Manga109} \\ 
\midrule
-&- & -  & - & PSNR$\uparrow$  / SSIM$\uparrow$  & PSNR$\uparrow$  / SSIM$\uparrow$  & PSNR$\uparrow$ / SSIM$\uparrow$  & PSNR$\uparrow$ / SSIM $\uparrow$ & PSNR$\uparrow$ / SSIM $\uparrow$ \\ \hline 
EDSR-B \cite{lim2017enhanced} & 1373K & 315.6G & $\times 2$ & 36.46 / 0.9604 &  34.49/ 0.9265 & 32.26 /  0.8996 &31.97 / 0.9271 & 38.45 / 0.9758 \\ \hline
CARN \cite{ahn2018fast} & 1587K & 221.7G & $\times 2$ & 36.86 /  0.9530 &  34.32 /  0.9176 & 32.19/  0.8987 &  31.82 / 0.9236 & 38.36 / 0.9765 \\ \hline
IMDN \cite{hui2019lightweight} &  692K & 158.4G& $\times 2$ &  38.02 /  0.9606 &  33.66 / 0.9179 & 32.24 / 0.8998 & 32.25 / 0.9285 & 38.25 / 0.9759\\ \hline
LatticeNet \cite{luo2020latticenet} & 751K &  168.2G & $\times 2$ & 38.08 / 0.9611 &33.72 / 0.9188 & 32.24 /  0.9010 & 32.35 /  0.9288 & 38.84 / 0.9754 \\ \hline
SwinIR-L \cite{liang2021swinir} & 911K & 244.5G & $\times 2$ & 38.12 /  0.9611&  33.77 /  0.9216 &  32.36 /  0.9018 & 32.75 / 0.9345 & 39.08 / 0.9781 \\ \hline
 SwinIR-NG\cite{choi2023n} &    1181K &  274.1G & $\times 2$ & 38.17 /  0.9612 &  33.94 / 0.9205 & 32.31 /   0.9013 &    32.78 /   0.9340 &    39.20 /    0.9781 \\ \hline
 SRFormer-L\cite{zhou2023srformer} & 856K &  237.5G & $\times 2$ & 38.19 / 0.9643&  33.92 / 0.9207 & 32.42 /  0.9025 & 32.96 /  0.9363 &39.29 / 0.9787\\ \hline
HiT-SIR\cite{aslahishahri2024hitsr} &847K& 226.5G & $\times 2$ & 38.25 /  0.9613 &  34.11 / 0.9216 & 32.39 /  0.9026 & 33.15 / 0.9381 &  39.51 /  0.9788 \\ \hline

\rowcolor[HTML]{FFC0CB} WaveHiT-SIR &  785K &  211.5G & $\times 2$ & 38.29 /  0.9617 & 34.14 / 0.9223 & {\color[HTML]{0070C0} \textbf{ 32.69 }}  / {\color[HTML]{0070C0} \textbf{ 0.9042 }}  & 33.22 / 0.9392 &  39.57 /  0.9791 \\ \hline
\rowcolor[HTML]{FFC0CB} WaveHiT-SNG &  1038K &  221.2G  & $\times 2$ &38.32 /  0.9619 & 34.17 / 0.9226 & 32.49 /  0.9035 & 33.29 / 0.9395 &  39.61 /  0.9795  \\ \hline
\rowcolor[HTML]{FFC0CB} WaveHiT-SRF &    832K & 216.2G &  $\times 2$ & {\color[HTML]{0070C0} \textbf{ 38.45 }}   / {\color[HTML]{0070C0} \textbf{ 0.9659 }}   & {\color[HTML]{0070C0} \textbf{  34.26 }} / {\color[HTML]{0070C0} \textbf{  0.9236 }}  &  32.46   /  0.9029      & {\color[HTML]{0070C0} \textbf{ 33.66 }}  / {\color[HTML]{0070C0} \textbf{ 0.9415 }}  & {\color[HTML]{0070C0} \textbf{ 39.77 }}  / {\color[HTML]{0070C0} \textbf{ 0.9825 }}    \\ \hline 
\\ 
\midrule
EDSR-B \cite{lim2017enhanced} & 1557K & 160.3G & $\times 3$ &  34.35 /  0.9268 &  30.31 /  0.8527 &29.19/ 0.8024 & 28.39 / 0.8544 &  33.55 /  0.9443 \\ \hline
CARN \cite{ahn2018fast} &  1594K & 118.9G & $\times 3$ & 34.25/ 0.9245 &  30.29 /  0.8407 & 29.07 / 0.8046 & 28.16 /  0.8496 & 33.52 / 0.9452 \\ \hline
IMDN \cite{hui2019lightweight} & 713K & 72.4G & $\times 3$ & 34.46 /  0.9280 &  30.32 / 0.8417 &  29.14 / .8056 & 28.23 / 0.8523 & 33.61 /  0.9445 \\ \hline
LatticeNet \cite{luo2020latticenet} &  774K &  76.7G & $\times 3$ & 34.42/  0.9284 &  30.41 / 0.8426 & 29.10 /  0.8049 &  28.20 / 0.8511 &  33.72 /  0.9444 \\ \hline
SwinIR-L \cite{liang2021swinir} &  918K & 110.8G & $\times 3$ &34.61 / 0.9289 & 30.53 / 0.8461 &  29.21 / 0.8084 &  28.68 /0.8626 & 34.01 /  0.9482 \\ \hline
  SwinIR-NG\cite{choi2023n} &   1190K& 114.1G& $\times 3$ & 34.64 /  0.9293 &  30.58 / 0.8471 &  29.24 /  0.8090 &   28.75 /  0.8639 &   34.22 /   0.9488 \\ \hline
  SRFormer-L\cite{zhou2023srformer} &  872K& 106.8G& $\times 3$ & 34.71 /  0.9297 & 30.58 / 0.8471 & 29.28 / 0.8099 &  28.82 / 0.8658 &  34.22 /  0.9489 \\ \hline
HiT-SIR\cite{aslahishahri2024hitsr} &  855K & 101.6G & $\times 3$ &  34.74 /  0.9301 & 30.62 /  0.8477 & 29.32 / 0.8117 & 29.02 /  0.8698 &  34.54 / 0.9504 \\ \hline
\rowcolor[HTML]{FFC0CB} WaveHiT-SIR &  774K &  93.2G & $\times 3$ & {\color[HTML]{0070C0} \textbf{ 34.81 }} /   0.9309  & {\color[HTML]{0070C0} \textbf{ 31.02 }} /  {\color[HTML]{0070C0} \textbf{ 0.8537 }} & 29.57 / 0.8214  & 29.38 /0.8858  & 34.74 / 0.9574 \\ \hline
\rowcolor[HTML]{FFC0CB} WaveHiT-SNG & 1043K &  101.4G & $\times 3$ &35.12 / {\color[HTML]{0070C0} \textbf{ 0.9420  }} & 30.82 /  0.8492 & 29.67 / 0.8222& 29.37 /0.8813 & \color[HTML]{0070C0} \textbf{ 34.84 }  / \color[HTML]{0070C0} \textbf{ 0.9584 }   \\ \hline
\rowcolor[HTML]{FFC0CB} WaveHiT-SRF &  843K& 103.7G & $\times 3$& 34.78 /  0.9306 & 30.90 /  0.8501 & \color[HTML]{0070C0} \textbf{ 29.87 } / \color[HTML]{0070C0} \textbf{ 0.8275 } & \color[HTML]{0070C0} \textbf{ 29.42 } / \color[HTML]{0070C0} \textbf{ 0.8898 } & 34.64 / 0.9544  \\ \hline 
\\ 
\midrule
EDSR-B \cite{lim2017enhanced} & 1526K & 114.2G & $\times 4$ & 32.07/ 0.8936 & 28.54 / 0.7809 & 27.55 /  0.7349 & 26.14 / 0.7856 & 30.34 /  0.9063 \\ \hline
CARN \cite{ahn2018fast} &  1594K &  91.5G & $\times 4$ &  32.11 / 0.8935 &  28.62 / 0.7819 & 27.59 /0.7352 &  26.09/ 0.7841 &  30.49 /  0.9086 \\ \hline
IMDN \cite{hui2019lightweight} & 716K & 41.7G& $\times 4$ &  32.20 / 0.8947 &  28.56 / 0.7813 &  27.57 / 0.7359 &  26.09 /  0.7845 & 30.49 /  0.9079 \\ \hline
LatticeNet \cite{luo2020latticenet} & 779K &  43.8G & $\times 4$ & 32.18 /  0.8943 & 28.64 /  0.7820 & 27.59 / 0.7361 & 26.14 / 0.7844 & 30.54 /  0.9076 \\ \hline
SwinIR-L \cite{liang2021swinir} &  931K &  63.8G & $\times 4$ & 32.45 / 0.8978 & 28.78 /  0.7859 & 27.70 / 0.7407 & 26.48 /0.7981 & 30.93/ 0.9152\\ \hline
 SwinIR-NG\cite{choi2023n} &  1201K &    64.4G & $\times 4$ &  32.44 /  0.8980 &  28.83 /   0.7870 &  27.73 /   0.7418 & 26.61 /   0.8010 &  31.09/  0.9161 \\ \hline
 SRFormer-L\cite{zhou2023srformer} &  875K&  63.9G & $\times 4$ &  32.51 /  0.8988 & 28.83 /  0.7874 & 27.73 /  0.7422 & 26.69 /  0.8035 & 31.18/ 0.9167 \\ \hline
HiT-SIR\cite{aslahishahri2024hitsr} &  869K & 58.2G & $\times 4$ & 32.55 /  0.8999 &  28.87 / 0.7880 & 27.75 / 0.7432 & 26.80/ 0.8069 &  31.26 /0.9171\\ \hline
\rowcolor[HTML]{FFC0CB} WaveHiT-SIR &  781K & 52.6G & $\times 4$ & 32.57 /  0.9003 &29.17 / 0.80050 & 27.88 / 0.7512 & 27.21 /0.8142 & \color[HTML]{0070C0} \textbf{ 31.87 } / \color[HTML]{0070C0} \textbf{ 0.9282 } \\ \hline
\rowcolor[HTML]{FFC0CB} WaveHiT-SNG &  1054K &  59.7G & $\times 4$ & 32.84 /  0.9023 & 29.13 / 0.7913  & \color[HTML]{0070C0} \textbf{ 27.93 }  / \color[HTML]{0070C0} \textbf{ 0.7562 } & 27.17 /0.8122 & 31.27 /0.9182  \\ \hline
\rowcolor[HTML]{FFC0CB} WaveHiT-SRF &  857K &  56.4G & $\times 4$ & \color[HTML]{0070C0} \textbf{ 32.98 }  /\color[HTML]{0070C0} \textbf{ 0.9053 }   & \color[HTML]{0070C0} \textbf{ 29.32 }   / \color[HTML]{0070C0} \textbf{ 0.80253 }  & 27.91 / 0.7522 & \color[HTML]{0070C0} \textbf{ 27.29 }  / \color[HTML]{0070C0} \textbf{ 0.8173 }  & 31.57 /0.9262   \\ 
\bottomrule
\end{tabular}}
\end{table*}

\subsection{Qualitative Results}

Figure \ref{fig:Urban100} shows qualitative comparisons for challenging image SR tasks. Traditional SR methods often produce blurred images and artifacts (e.g. $img095$) due to their focus on local features and small receptive fields. In contrast, WaveHiT-SIR and WaveHiT-SRF can capture long-range dependencies with larger windows, resulting in sharper details and more refined textures.
Structure distortion is a typical problem in SR, particularly with complex images such as $img077$. Traditional methods struggle to restore image structure, resulting in blurred lines. WaveHiT-SR, however, leverages multiscale features to preserve structure and enhance detail. Models like WaveHiT-SIR and WaveHiT-SRF provide superior image structure restoration compared to SwinIR-Light \cite{liang2021swinir} and SRFormer-Light \cite{zhou2023srformer}.

\subsection{Quantitative Results}
\begin{figure}[h]
\centering
\includegraphics[width=0.9\linewidth]{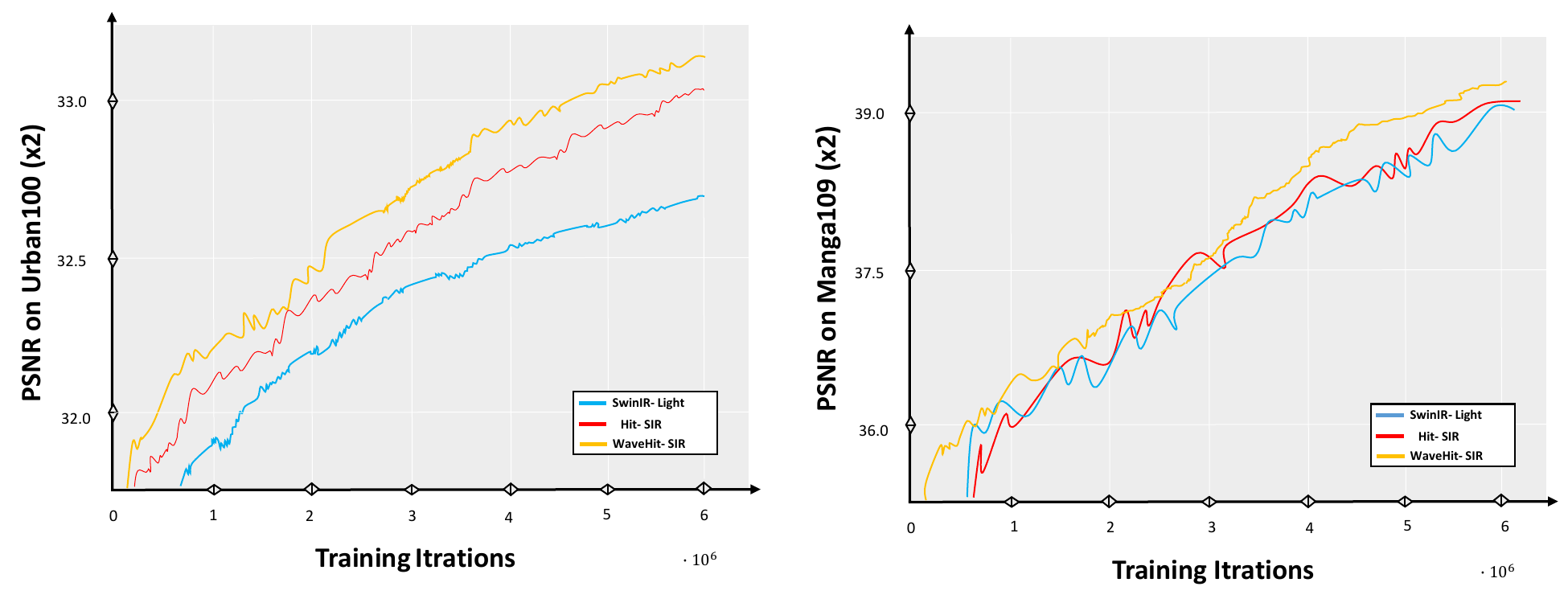}
\caption{To compare convergence rates, we evaluated SwinIR-Light \cite{liang2021swinir}, HIT-SR \cite{aslahishahri2024hitsr}, and our optimized model, WaveHiT-SR, using the Urban100 $(\times2)$ and Manga109 $(\times2)$ datasets. All models were trained under consistent conditions, enabling a fair assessment of convergence behavior and performance improvements.}
\label{Training}
\end{figure}

As shown in Table \ref{tab:PSNR and SSIM}, our WaveHiT-SR methods achieve outstanding results in all benchmark datasets. Compared to existing state-of-the-art models such as EDSR-B \cite{lim2017enhanced}, CARN \cite{ahn2018fast}, IMDN \cite{hui2019lightweight}, LatticeNet \cite{luo2020latticenet}, SwinIR-L \cite{liang2021swinir}, SwinIR-NG \cite{choi2023n},  SRFormer-Light \cite{zhou2023srformer}, and HiT-SR \cite{aslahishahri2024hitsr}, our proposed WaveHiT-SIR method delivers superior qualitative and quantitative performance while reducing both the model size and computational demands. Our updated WaveHiT-SRF models outperform the original versions in terms of performance, efficiency, and convergence. Table \ref{tab:PSNR and SSIM} illustrates the PSNR improvements our methods (WaveHiT-SIR, WaveHiT-SNG, WaveHiT-SRF) achieve, with gains of \textbf{0.49}, \textbf{0.41}, and \textbf{0.48} dB on Urban100 $(\times2)$. Replacing the inefficient shifted window self-attention with WaveAttention-Channel Correlation (WA-SCC) results in lower computational demands and higher speed-up inference for all WaveHiT-SR models. Despite smaller FLOP and parameter improvements, WaveHiT-SRF outperforms SRFormer-Light in SR quality and offers a $6\times$ inference speed boost, benefiting practical applications.  Figure \ref{Training} compares the convergence of SwinIR-Light and WaveHiT-SIR on Urban100 $(\times2)$ and Manga109 $(\times2)$.
\vspace{-5pt}
\vspace{-8pt}

\section{Conclusion}

In this paper, we introduce WaveHiT-SR, a method that integrates wavelet transforms into a hierarchical transformer framework to advance image super-resolution (SR). Our strategy involves adapting popular transformer-based SR models to hierarchical transformers, enhancing SR efficiency (WaveHiT-SR). Our method incorporates WaveAttention, instead of spatial attention technique leveraging wavelet transforms to capture essential low- and high-frequency details, and employs both block-level and layer-level architectures. Each transformer block utilizes expanding hierarchical windows to capture long-range dependencies and multi-scale features, leading to improved SR outcomes. To address the high computational demands of self-attention, we introduce the WA-SCC, achieving linear complexity relative to window sizes for efficient hierarchical feature aggregation. This approach allows the model to focus on low-frequency details and enhance high-frequency textures, striking a balance between computational complexity and SR performance. 


\bibliography{aaai25}

\end{document}